\documentclass[10pt,twocolumn,letterpaper]{article}

\usepackage[pagenumbers]{cvpr} % To force page numbers, e.g. for an arXiv version

\usepackage{graphicx}
\usepackage{amsmath}
\usepackage{amssymb}
\usepackage{booktabs}
\usepackage{paralist}

\usepackage[pagebackref,breaklinks,colorlinks]{hyperref}

\usepackage[capitalize]{cleveref}
\crefname{section}{Sec.}{Secs.}
\Crefname{section}{Section}{Sections}
\Crefname{table}{Table}{Tables}
\crefname{table}{Tab.}{Tabs.}

\def\cvprPaperID{5910} % *** Enter the CVPR Paper ID here
\def\confName{CVPR}
\def\confYear{2022}

\begin{document}

%%%%%%%%% TITLE - PLEASE UPDATE
\title{Dyadic Human Motion Prediction}

\author{
	Isinsu Katircioglu$^1$
	\and
	Costa Georgantas$^2$
	\and
	Mathieu Salzmann$^{1,3}$
	\and
	Pascal Fua$^1$   
	\and
	$^1$CVLab, EPFL, Switzerland\quad
	$^2$CHUV, Switzerland\\
	$^3$ClearSpace SA, Switzerland
}

\maketitle
% !TEX root = ../top.tex
% !TEX spellcheck = en-US

\newif\ifdraft
%\drafttrue
\draftfalse
\ifdraft
\newcommand{\PF}[1]{{\color{red}{\bf PF: #1}}}
\newcommand{\pf}[1]{{\color{red} #1}}
\newcommand{\IK}[1]{{\color{blue}{\bf IK: #1}}}
\newcommand{\ik}[1]{{\color{blue} #1}}
\newcommand{\HR}[1]{{\color{magenta}{\bf hr: #1}}}
\newcommand{\hr}[1]{{\color{magenta} #1}}
\newcommand{\VC}[1]{{\color{blue}{\bf vc: #1}}}
\newcommand{\vc}[1]{{\color{blue} #1}}
\newcommand{\ms}[1]{{\color{green}{#1}}}
\newcommand{\MS}[1]{{\color{green}{\bf ms: #1}}}
\newcommand{\JS}[1]{{\color{cyan}{\bf js: #1}}}
\newcommand{\NEW}[1]{{\color{red}{#1}}}

\else
\newcommand{\PF}[1]{{\color{red}{}}}
\newcommand{\pf}[1]{ #1 }
\newcommand{\HR}[1]{{\color{blue}{}}}
\newcommand{\hr}[1]{#1}%
\newcommand{\VC}[1]{{\color{green}{}}}
\newcommand{\ms}[1]{ #1 }
\newcommand{\MS}[1]{{\color{green}{}}}
\newcommand{\NEW}[1]{#1}
\fi

\newcommand{\comment}[1]{}

\newcommand{\va}{\mathbf{a}}
\newcommand{\vb}{\mathbf{b}}
\newcommand{\vcc}{\mathbf{c}}
\newcommand{\vd}{\mathbf{d}}
\newcommand{\ve}{\mathbf{e}}
\newcommand{\vf}{\mathbf{f}}
\newcommand{\vg}{\mathbf{g}}
\newcommand{\vh}{\mathbf{h}}
\newcommand{\vi}{\mathbf{i}}
\newcommand{\vj}{\mathbf{j}}
\newcommand{\vk}{\mathbf{k}}
\newcommand{\vl}{\mathbf{l}}
\newcommand{\vm}{\mathbf{m}}
\newcommand{\vn}{\mathbf{n}}
\newcommand{\vo}{\mathbf{o}}
\newcommand{\vp}{\mathbf{p}}
\newcommand{\vq}{\mathbf{q}}
\newcommand{\vr}{\mathbf{r}}
\newcommand{\vt}{\mathbf{t}}
\newcommand{\vu}{\mathbf{u}}
\newcommand{\vv}{\mathbf{v}}
\newcommand{\vw}{\mathbf{w}}
\newcommand{\vx}{\mathbf{x}}
\newcommand{\vy}{\mathbf{y}}
\newcommand{\vz}{\mathbf{z}}

\newcommand{\mA}{\mathbf{A}}
\newcommand{\mB}{\mathbf{B}}
\newcommand{\mC}{\mathbf{C}}
\newcommand{\mD}{\mathbf{D}}
\newcommand{\mE}{\mathbf{E}}
\newcommand{\mF}{\mathbf{F}}
\newcommand{\mG}{\mathbf{G}}
\newcommand{\mH}{\mathbf{H}}
\newcommand{\mI}{\mathbf{I}}
\newcommand{\mJ}{\mathbf{J}}
\newcommand{\mK}{\mathbf{K}}
\newcommand{\mL}{\mathbf{L}}
\newcommand{\mM}{\mathbf{M}}
\newcommand{\mN}{\mathbf{N}}
\newcommand{\mO}{\mathbf{O}}
\newcommand{\mP}{\mathbf{P}}
\newcommand{\mQ}{\mathbf{Q}}
\newcommand{\mR}{\mathbf{R}}
\newcommand{\mS}{\mathbf{S}}
\newcommand{\mT}{\mathbf{T}}
\newcommand{\mU}{\mathbf{U}}
\newcommand{\mV}{\mathbf{V}}
\newcommand{\mW}{\mathbf{W}}
\newcommand{\mX}{\mathbf{X}}
\newcommand{\mY}{\mathbf{Y}}
\newcommand{\mZ}{\mathbf{Z}}

\newcommand{\cA}{\mathcal A}
\newcommand{\cB}{\mathcal B}
\newcommand{\cC}{\mathcal C}
\newcommand{\cD}{\mathcal D}
\newcommand{\cE}{\mathcal E}
\newcommand{\cF}{\mathcal F}
\newcommand{\cG}{\mathcal G}
\newcommand{\cH}{\mathcal H}
\newcommand{\cI}{\mathcal I}
\newcommand{\cJ}{\mathcal J}
\newcommand{\cK}{\mathcal K}
\newcommand{\cL}{\mathcal L}
\newcommand{\cM}{\mathcal M}
\newcommand{\cN}{\mathcal N}
\newcommand{\cO}{\mathcal O}
\newcommand{\cP}{\mathcal P}
\newcommand{\cQ}{\mathcal Q}
\newcommand{\cR}{\mathcal R}
\newcommand{\cS}{\mathcal S}
\newcommand{\cT}{\mathcal T}
\newcommand{\cU}{\mathcal U}
\newcommand{\cV}{\mathcal V}
\newcommand{\cW}{\mathcal W}
\newcommand{\cX}{\mathcal X}
\newcommand{\cY}{\mathcal Y}
\newcommand{\cZ}{\mathcal Z}

\newcommand*\rot{\rotatebox{90}}
\newcommand*\OK{\ding{51}}

\newcommand{\TODO}[1]{\textcolor{cyan}{#1}}
\definecolor{gray}{RGB}{127,127,127}
\newcommand{\optional}[1]{\textcolor{gray}{#1}}

\newcommand{\ST}{\mathcal{T}}
\newcommand{\SST}{\mathcal{T}_S}

\newcommand{\R}{\mathbb{R}}
\newcommand{\Seg}{\mathbf{S}} % geometric part
\newcommand{\Latent}{\mathbf{L}}
\newcommand{\LatentG}{\Latent^{\text{3D}}} % geometric part
\newcommand{\LatentA}{\Latent^\text{app}} % appearance part
\newcommand{\LatentBG}{\mB} % appearance part

\newcommand{\loss}{L}
\newcommand{\objFG}{O}
\newcommand{\objBG}{G}

\newcommand{\norm}[1]{\left\lVert#1\right\rVert}
\newcommand{\argmin}{\operatornamewithlimits{argmin}}
\newcommand{\erf}{\operatornamewithlimits{erf}}

\newcommand{\Var}{\operatornamewithlimits{Var}}

\newcommand{\parag}[1]{\vspace{-3mm}\paragraph{#1}}

\newcommand{\handheld}[0]{{\it Handheld190k}}
\newcommand{\ski}[0]{{\it Ski-PTZ-Dataset}}
\newcommand{\human}[0]{{\it H36M}}
\newcommand{\lindyhop}[0]{{\it LindyHop600K}}

\newcommand{\ours}[0]{{\bf Ours}}
\newcommand{\direct}[0]{{\bf Resnet}}
\newcommand{\LCR}[0]{{\bf LCR}}
\newcommand{\ECCV}[0]{{\bf NVS-encoder}}
\newcommand{\CVPR}[0]{{\bf Multiview}}
\newcommand{\auto}[0]{{\bf Auto-encoder}}

\begin{abstract}
Prior work on human motion forecasting has mostly focused on predicting the future motion of single subjects in isolation from their past pose sequence. In the presence of closely interacting people, however, this strategy fails to account for the dependencies between the different subject's motions. In this paper, we therefore introduce a motion prediction framework that explicitly reasons about the interactions of two observed subjects. Specifically, we achieve this by introducing a pairwise attention mechanism that models the mutual dependencies in the motion history of the two subjects. This allows us to preserve the long-term motion dynamics in a more realistic way and more robustly predict unusual and fast-paced movements, such as the ones occurring in a dance scenario. To evaluate this, and because no existing motion prediction datasets depict two closely-interacting subjects, we introduce the  LindyHop600K dance dataset. Our results evidence that our approach outperforms the state-of-the-art single person motion prediction techniques. 
\end{abstract}

\section{Introduction}

Forecasting future motion from observed past 3D poses has primarily been studied in a single-person setting~\cite{Li18k,Mao19,Mao20,Lebailly20,Lingwei21}. A naive way to extend these approaches to the multi-person case is to simply treat each subject independently. However, this fails to account for interactions that condition future behavior. Only in~\cite{Adeli20} is there an attempt to capture them via the use of social cues obtained by pooling the learned features  for each individual. While effective in the presence of weak social interactions, this approach is ill-suited to modeling the stronger dependencies that arise from two closely-interacting individuals whose movements are highly correlated.

In this paper, we therefore introduce an approach to dyadic, or pairwise, human motion prediction that more strongly models interactions. To this end, we develop an encoder-decoder architecture with both self- and pairwise attention modules. While self-attention captures the similarities between someone's present and past motions, pairwise attention models capture the dependencies between the pose histories of both subjects. Then, for each subject, the decoder takes as input the subject's own self-attention features and the pairwise attention ones, and outputs the future 3D pose sequence.

As there is no dyadic motion prediction benchmark with closely-interacting people, we build the \lindyhop{} dataset. It features Lindy Hop dancers performing  energetic moves, ranging from frenzied kicks to smooth and sophisticated body motions. The dancers synchronize their fast-paced steps with one another and the music. The standard footwork can be followed by infrequent twirls, which make the upcoming pose prediction hard without observing the highly correlated moves of the partner. The motion of one person gives significant clues about infrequent or subtle motion patterns of the other that cannot be easily inferred from the isolated individual motion.

Hence, our contributions are twofold.
\begin{compactitem} 

	\item We propose the first 3D motion prediction method that models the dyadic motion dependencies between two subjects.
	 
	\item We introduce a new dance dataset, \lindyhop{}, which consists of videos and 3D human body poses of dancers performing diverse swing motions.
	
\end{compactitem}
Our experiments on the \lindyhop{} dataset clearly demonstrate the benefits of our method. It outperforms both the state-of-the-art single person baselines and the use of weaker social cues~\cite{Adeli20}. Our results are especially promising in terms of long-term prediction. The proposed method models the motion dynamics much more reliably than the baselines. Our code and dataset will be made publicly available.

\section{Related Work}

\subsection{Single-actor Motion Prediction} Although early approaches to human motion prediction relied on traditional models, such as Hidden Markov Models~\cite{Brand00}, Gaussian Processes for time-series analysis~\cite{Wang05b}, conditional restricted Boltzmann machine~\cite{Taylor06} and dynamic random forest~\cite{Lehrmann14}, the state-of-the-art in this field is now achieved via deep networks.
As motion forecasting inherently is a sequence-to-sequence prediction task~\cite{Sutskever11,Sutskever14,Bahdanau15,Dutil17,Vaswani17}, much work has focused on encoder-decoder models. This was pioneered by the Encoder-Recurrent-Decoder (ERD) model of~\cite{Fragkiadaki15}, which led to the development of several RNN-based strategies. For example,~\cite{Jain16a} introduces a Structural-RNN (S-RNN) based on a spatio-temporal graph that encapsulates the dependencies among the body joints over time; \cite{Ghosh17} leverages de-noising autoencoders to learn the spatial structure of the human skeleton by randomly removing information about joints during training; \cite{Martinez17b,Chiu19b} use velocities instead of poses; \cite{Gopalakrishnan19} similarly integrates motion derivatives; \cite{Zhou18a} uses an auto-regressive approach for long-term prediction. Unfortunately, these RNN-based methods tend to produce discontinuities at the transition between the last observed pose and the first predicted one.

To address this limitation, several methods have attempted to better model the distribution of valid motions. In this context, \cite{Gui18a} integrates adversarial training to enforce frame-wise geometric plausibility and sequence-wise coherence; \cite{Ruiz19} uses a GAN with several discriminators that operates on input sequences with masked joints and learns to inpaint the missing information; \cite{Cui21} exploits a similar GAN but uses spectral normalization to perform temporal attention; \cite{Wang19h} formulates motion prediction as a generative adversarial imitation learning task to focus on shorter sequences by breaking long ones into small chunks. In contrast to the previous methods that generate a single future prediction, several GAN- and VAE-based methods aim to produce multiple diverse future motion sequences~\cite{Barsoum18,Yan18a,Kundu19,Yuan20,Aliakbarian20,Aliakbarian21,Mao21b}. For example, \cite{Aliakbarian20,Aliakbarian21} achieve this via a conditional variational autoencoder (CVAE); \cite{Mao21b} generates the motion of different body parts sequentially; \cite{Yuan20} proposes a novel sampling strategy to produce diverse samples from a pretrained generative model; \cite{Cao20,Wang21d} generate multiple trajectory and pose predictions conditioned on the scene context. While these methods produce smoother transitions than RNNs, they do not explicitly model the dependencies between the different body joints.

Recently, this has been addressed via graph convolutional networks (GCNs). In particular, \cite{Mao19} encodes the spatio-temporal relationships across the joints via a GCN that adaptively learns the body connectivity; \cite{Lebailly20} also relies on a GCN and processes the past sequence at different lengths; \cite{Lingwei21} proposes to predict the poses first at a coarse level, and then at finer levels using a multi-scale residual GCN; Similarly,~\cite{Li21b} employs a multi-scale GCN that jointly learns action categories and motion dynamics at different granularities.

As an alternative to RNNs, GANs and VAEs, several methods rely on attention-based models, which proved to be effective in machine translation and image caption generation~\cite{Bahdanau15, Xu15, Vaswani17}. In particular, \cite{Tang18c} proposes an attention mechanism to focus on the moving joints of the human body for motion forecasting; \cite{Shu20} uses a similar idea to learn the spatial coherence and temporal evolution of joints via a co-attention mechanism; \cite{Mao20} combines a GCN with an attention module to learn the repetitive motion patterns from the past;  \cite{Mao21a} fuses the predictions from three attention modules that process motion at different levels: full body, body parts, and individual joints; \cite{Gonzalez21} trains a computationally less intensive Transformer~\cite{Vaswani17} to infer the future poses in parallel.

In any event, all the above-mentioned methods tackle the single-actor scenario. As such, and as will be evidenced by our experiments, they are sub-optimal to handle the case of two closely-interacting subjects.

\subsection{Social Interactions in Motion Prediction}
Modeling human-to-human interactions is a long-studied problem, with much focus on the social dynamics occurring in a group of people~\cite{Helbing95}. In particular, much work in this space has been dedicated to the problem of trajectory prediction~\cite{Alahi14,Mehran09,Pellegrini09,Yamaguchi11,Robicquet16}, whose goal is to predict the global motion of people in a group, not their detailed 3D pose. In this context, recent methods have also studied the use of RNNs~\cite{Alahi16,Santaro17a,Deo18,Sun18c,Sun19c}, GANs~\cite{Gupta18,Sadeghian19,Kosaraju19}, graph neural networks~\cite{Huang19b,Casas20,Zhang21}, and attention mechanisms~\cite{Li21c,Li21d,Liu21c,Shafiee21}, but also of reinforcement~\cite{Lee17a} and contrastive~\cite{Liu21d} learning.

Another task that has benefited from modeling social interactions is 3D multi-person pose estimation~\cite{Corona20a,Guo21,Wang20f,Ng20}. However, the scenarios studied in this context typically do not involve closely-interacting subjects, but rather individuals exchanging objects, crossing path, or coexisting in an environment during a short period of time. Therefore, the existing solutions only aim to encode weak constraints arising from such social interactions instead of exploiting strong dependencies as we do here.

To the best of our knowledge,~\cite{Adeli20,Guo21b} are the only works that, as us, target multi-person motion prediction.~\cite{Adeli20} uses a social pooling layer to fuse the features corresponding to the encoded past motion of each subject. However, by relying on either $\textit{max}$, $\textit{average}$ or $\textit{sum}$ pooling of the individuals' features, it only encodes weak dependencies between the subjects and was demonstrated in scenarios where the motions of the individuals are only weakly correlated. By contrast, we focus on the case of two closely-interacting subjects, and introduce an approach that models the subjects' dependencies. Concurrently,~\cite{Guo21b} has also been working on multi-person motion prediction for a lindy hop scenario. However, unlike our method,~\cite{Guo21b} is not agnostic to the roles of the dancers and requires separate pipelines and losses based on the role. In addition to that~\cite{Guo21b} does not exploit relative motion between the interacting people which is an integral part of our model. Since the training code and dataset of this work are not publicly available, we could not compare against them. 

\label{sec:related}

\section{Method}

Let us now introduce dyadic human motion prediction method for closely-interacting people. To this end, we first review the single person motion prediction formalism at the heart of our method, and then present our approach to modeling pairwise interactions to predict the future poses of two people. 

\subsection{Single Person Baseline}
\begin{figure}[t]
	\centering
	\begin{tabular}{c}
		\includegraphics[width=0.6\linewidth]{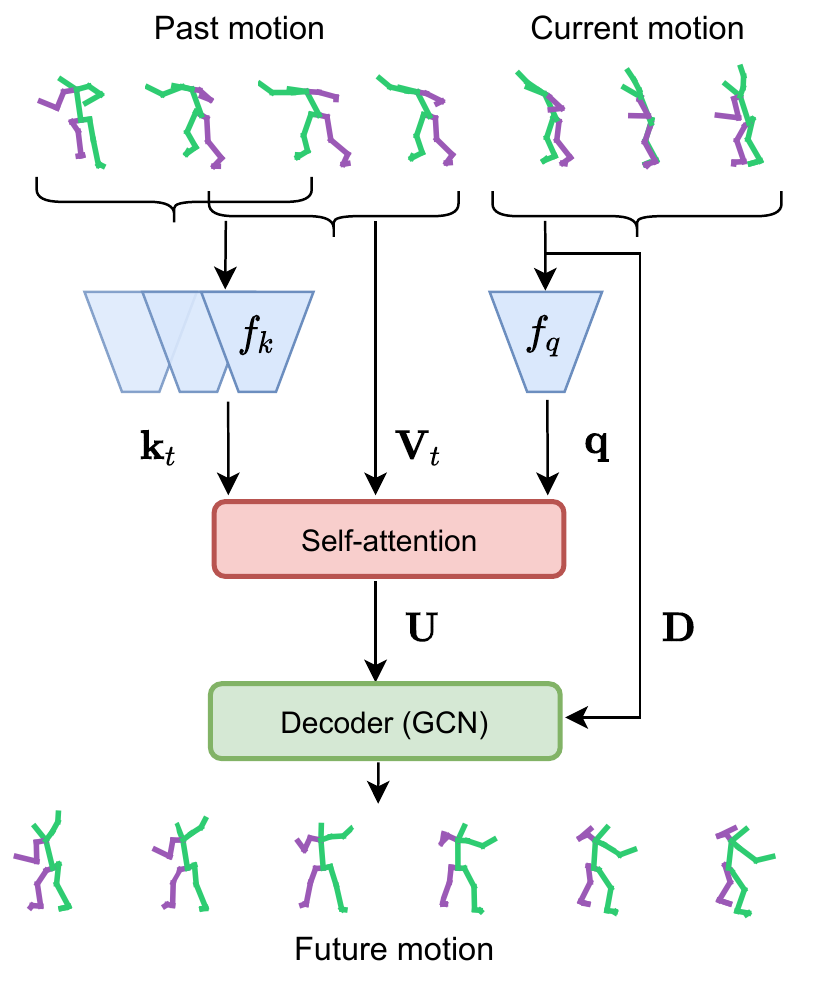} \\
	\end{tabular}
	\caption[Single person motion forecasting baseline]{\textbf{Single person motion forecasting baseline~\cite{Mao20}.} The baseline model aggregates information from the history of poses (keys) by comparing them to the last observed sequence of poses (query) through an attention mechanism. $f_k$ and $f_q$ are modeled with convolutional layers. The weighted sum of the values are concatenated with the DCT coefficients of the last observed poses and fed into the GCN that outputs the future pose predictions.}
	\label{fig:baseline_3dmotion_forecasting}
\end{figure}
Our work builds on ``History Repeats Itself (HRI)"~\cite{Mao20}, which relies on an attention mechanism and a GCN to predict the future poses of a single person based on their observed sequence of historical poses. Intuitively, the attention mechanism aims to focus the prediction on the most relevant parts of the motion history and the GCN decodes the resulting representation into the future pose predictions while encoding the dependencies across the different joints. 

Formally, given a sequence of ${T}_p$ past 3D poses of an individual, $\textbf{X}_{1:{T}_p}=[\textbf{x}_{1}, \textbf{x}_{2}, ...,\textbf{x}_{{T}_p}]^T$, single-person human motion prediction aims to estimate the ${T}_f$ future 3D poses $\textbf{X}_{{T}_p+1:{T}_p+{T}_f}=[\textbf{x}_{{T}_p+1}, \textbf{x}_{{T}_p+2}, ...,\textbf{x}_{{T}_p+{T}_f}]^T$. Each pose $\textbf{x}_t \in \mathbb{R}^K$, where $K= J\times 3$, comprises $J$ joints forming a skeleton.
In HRI, the similarity between past motions and the last observed motion context is captured by dividing the motion history $\textbf{X}_{1:{T}_p}$ into ${T}_p-{T}_l-{T}_f+1$ sub-sequences $\{\textbf{X}_{t:t+{T}_l+{T}_f-1}\}_{t=1}^{{T}_p-{T}_l-{T}_f+1}$, each containing ${T}_l+{T}_f$ consecutive poses. The attention mechanism is then built by treating the first $T_l$ poses of every sub-sequence as key and the entire sub-sequence $\{\textbf{X}_{t:t+{T}_l+{T}_f-1}\}$ as value. In practice, the values are in fact represented in trajectory space as the Discrete Cosine Transform (DCT) coefficients of the corresponding poses. That is, the value of each subsequence is taken as $\{\textbf{V}_{t}\}_{t=1}^{{T}_p-{T}_l-{T}_f+1}$, where $\textbf{V}_{t} \in \mathbb{R}^{K \times ({T}_l+{T}_f)}$ encodes the DCT coefficients. Finally, the query corresponds to the last observed sub-sequence $\textbf{X}_{{T}_p-{T}_l+1:{T}_p}$ with $T_l$ poses. 

The query and keys are computed as the output of two neural networks $f_q$ and $f_k$, respectively. These functions map the poses to latent vectors of dimension $d$, that is, 
\begin{align}
	\textbf{q} &= f_q(\textbf{X}_{{T}_p-{T}_l+1:{T}_p}) \; , \label{eq:query_computation}\\
	\textbf{k}_{t} &= f_k(\textbf{X}_{t:t+{T}_l-1}) \; , \label{eq:key_computation}
\end{align}
where $\textbf{q}, \textbf{k}_{t} \in \mathbb{R}^d$ and $1 \leq t \leq {T}_p-{T}_l-{T}_f+1$.
A similarity score $a_{t}$ is then computed for each key-query pair, and these scores are employed to obtain a weighted combination of the values. This is expressed as
\begin{align}
	a_{t} = \frac{\textbf{q}\textbf{k}_{t}^T}{\sum_{j=1}^{{T}_p-{T}_l-{T}_f+1}\textbf{q}\textbf{k}_{j}^T} \; , \hspace{4mm}
	\textbf{U} = \sum_{t=1}^{{T}_p-{T}_l-{T}_f+1}a_{t} \textbf{V}_{t}, \label{eq:self_att}
\end{align}
where $\textbf{U} \in \mathbb{R}^{K \times ({T}_l+{T}_f)}$.
Then, the last observed sub-sequence is extended to a sequence of length ${T}_l+{T}_f$ by replicating the last pose and passed to the DCT module yielding $\textbf{D} \in \mathbb{R}^{K \times ({T}_l+{T}_f)}$. Finally, $\textbf{U}$ and $\textbf{D}$ are fed into the decoder GCN module, which outputs the future pose predictions $\hat{\textbf{X}}_{{T}_p+1:{T}_p+{T}_f}$. The attention module explained in this section is depicted in Fig.~\ref{fig:baseline_3dmotion_forecasting}, and will be referred to as self-attention in the rest of this paper, as it computes the attention of a single person on themselves.

\subsection{Pairwise Attention for Dyadic Interactions}

\begin{figure*}[t]
	\centering
	\begin{tabular}{c}
		\includegraphics[width=0.68\linewidth]{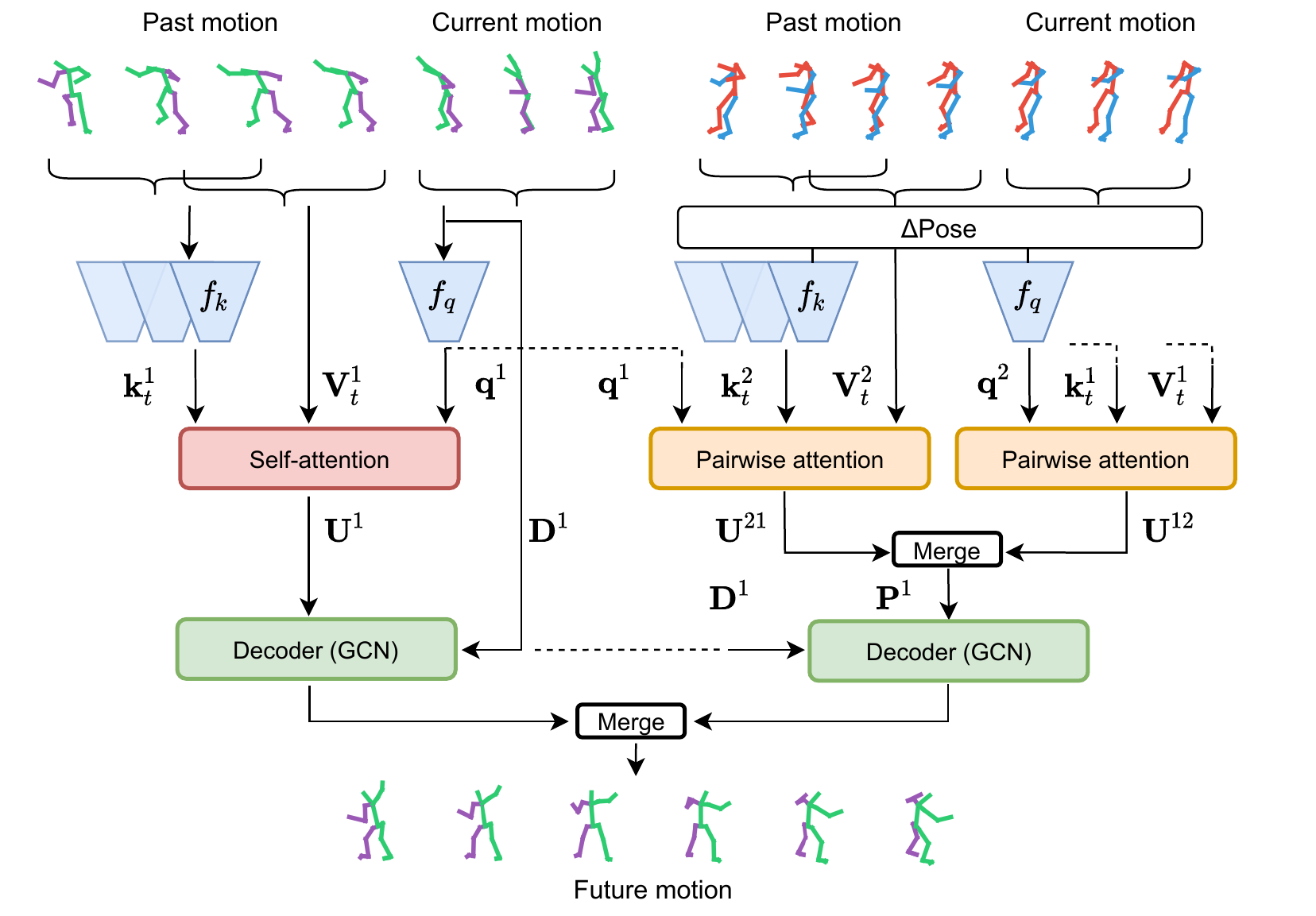} \\
	\end{tabular}
	\vspace{-3mm}
	\caption[Overview of our 3D motion forecasting model based on self- and pairwise attention]{\textbf{Overview of our 3D motion forecasting model based on self- and pairwise attention.} Our model takes as input the past poses of the primary (skeleton models depicted using green-purple) and past poses of the auxiliary (red-blue) subject relative to the primary one depicted by $\Delta \text{Pose}$ operation. The superscript $1$ is used for the primary subject whereas $2$ represents the interactee. As proposed by~\cite{Mao20}, the self-attention module takes as input the key, query and value vectors of the primary subject. We build on top of this approach by integrating a pairwise module that takes as input the query from one subject and key-value pair from the other subject. This module learns to put higher attention on the sub-sequences in the motion history of the primary subject that are more relevant to the current motion of the interactee. The merge block applies concatenation followed by a convolutional layer. The embeddings from self- and pairwise attention are fed into two separate GCNs with shared weights. The outputs of GCNs are projected to the future pose predictions of the primary subject via the merge block.}
	\label{fig:overview_3dmotion_forecasting}
	\vspace{-3mm}
\end{figure*}

Our goal is to perform motion predictions for multiple people. Formally, given the history of poses $\{\textbf{X}^{s}_{1:{T}_p}\}_{s=1}^{S}$ for $S$ subjects, our model predicts the future poses $\{\textbf{X}^{s}_{{T}_p+1:{T}_p+{T}_f}\}_{s=1}^{S}$. In particular, we focus on the case where $S = 2$ and aim to model the strong dependencies arising from the close interaction of the two subjects. As shown in Fig.~\ref{fig:overview_3dmotion_forecasting}, our approach combines self- and pairwise attention modules, and we refer to one person as the \emph{primary} subject and to the other as the \emph{auxiliary} one, denoted by the superscripts $1$ and $2$, respectively. Our goal then is to predict the future poses of the primary subject given the observed motions of both. Note that, to predict the future poses of the second subject, we simply inverse the roles.

To combine self- and pairwise attention, we first compute keys $\textbf{k}^{1}_{t}$ and query $\textbf{q}^{1}$ vectors for the primary subject as in Eqs.~\ref{eq:query_computation},~\ref{eq:key_computation}. The values $\textbf{V}^{1}_{t}$ together with $\textbf{k}^{1}_{t}$ and $\textbf{q}^{1}$ are then fed into the self-attention module, which yields ${\textbf{U}}^{1}$ as in Eq.~\ref{eq:self_att}. We then design a pairwise attention module that computes the similarity scores between the keys of the primary subject and the query of the auxiliary one, and vice-versa, to detect how relevant the coupled motion is at a given time in the past. A straightforward way of incorporating pairwise attention would consist of computing the auxiliary keys and query vectors directly from the observed motion of the auxiliary subject. However, as we show in the experiments, using the relative motion between the primary and auxiliary subject facilitates the modeling of interactions. Therefore, we compute the query, keys and values for the auxiliary subject as
\begin{align}
	\textbf{q}^{2} &= f_q(\textbf{X}^{1}_{{T}_p-{T}_l+1:{T}_p} - \textbf{X}^{2}_{{T}_p-{T}_l+1:{T}_p})  \; , \label{eq:pwise_query_computation}\\
	\textbf{k}^{2}_{t} &= f_k(\textbf{X}^{1}_{t:t+{T}_l-1} - \textbf{X}^{2}_{t:t+{T}_l-1}) \; , \label{eq:pwise_key_computation} \\
	\textbf{V}^{2}_{t}&= DCT(\textbf{X}^{1}_{t:t+{T}_l+{T}_f-1} - \textbf{X}^{2}_{t:t+{T}_l+{T}_f-1}) \; .\label{eq:pwise_value_computation}
\end{align}
We then define pairwise attention scores between the past motion of the primary subject and the relative motion with respect to the auxiliary one as
\begin{align}
	c^{12}_{t} = \frac{\textbf{q}^{2}{\textbf{k}^{1}}_{t}^T}{\sum_{j=1}^{{T}_p-{T}_l-{T}_f+1}\textbf{q}^{2}{\textbf{k}^{1}}_{j}^T} \; . \label{eq:pwise_scores}
\end{align}
This lets us compute a weighted sum of primary subject values as
\begin{align}
	\textbf{U}^{12} = \sum_{t=1}^{{T}_p-{T}_l-{T}_f+1}c^{12}_{t} \textbf{V}^{1}_{t} \;.
\end{align}
We also compute $\textbf{U}^{21}$ using $\textbf{V}^{2}_{t}$ and the pairwise scores  $c^{21}_{t}$ of $\textbf{q}^{1}$ and $\textbf{k}^{2}_{t}$. In the final stage of the encoder, we concatenate the pairwise embeddings ${\textbf{U}}^{12}$ and ${\textbf{U}}^{21}$ and feed them to a convolutional layer corresponding to the merge block in Fig.~\ref{fig:overview_3dmotion_forecasting}. The output is denoted as ${\textbf{P}}^{1}$. 

As for single-person prediction, the last observed sub-sequence of the primary subject is extended by repeating its last observed pose and transformed into DCT coefficients denoted by $\textbf{D}^{1}$. Our decoder then has two GCNs with shared parameters. One takes as input the concatenated matrices $\textbf{D}^{1}$ and $\textbf{U}^{1}$ and the other $\textbf{D}^{1}$ and $\textbf{P}^{1}$. Finally, the GCNs' outputs are projected via a convolutional layer to the future pose predictions of the primary subject. The same strategy is applied when exchanging the roles to obtain the future poses of the second subject.

\subsection{Training}
The entire network is trained by minimizing the Mean Per Joint Position Error (MPJPE). The loss for one training sequence is thus written as
\begin{equation}
\begin{split}
\textit{L} = \frac{1}{J({T}_l+{T}_f)}\sum_{t={T}_p-{T}_l+1}^{{T}_p+{T}_f}\sum_{j=1}^{J} ||\hat{\textbf{x}}_{t, j} - \textbf{x}_{t, j}||^{2}\;,
\end{split}
\end{equation}
where $\hat{\textbf{x}}_{t} \in \mathbb{R}^{3 \times J}$ encodes the estimated 3D pose for time $t$, $\textbf{x}_{t}$ represents the corresponding ground-truth pose, and $\textbf{x}_{t,j}$ denotes the 3D position of the $j$-th joint. 

\subsection{Implementation Details}

\parag{Training Details.} We train our network using the ADAM~\cite{Kingma15} optimizer with a learning rate of $0.0005$ and a batch size of $32$. We use $T_p = 60$ poses, corresponding to 2 seconds, as motion history  and predict $T_f = 30$ poses, corresponding to 1 second in the future. Our models are trained for 500 epochs, and we report the results of the model with the highest validation score.

\parag{Network Structure.} The networks $f_q$ and $f_k$ in the self- and pairwise attention modules consist of two 1D convolutional layers with kernel sizes $6$ and $5$, respectively, each followed by a ReLU. The hidden dimension of the query and key vectors in Eq.~\ref{eq:query_computation} and Eq.~\ref{eq:key_computation} is $256$. We use a GCN with $12$ residual blocks as in~\cite{Mao20}. The human skeleton has $J=19$ joints and our model has approximately $3.27M$ parameters similar to~\cite{Mao20} that has $3.26M$ parameters.

\section{Experiments}

In this section, we demonstrate the effectiveness of our approach at exploiting dyadic interactions. To this end, we first introduce our \lindyhop{} dataset depicting couples that perform lindy hop dance movements.

\subsection{LindyHop600K}
Lindy hop is a type of swing dance with fast-paced steps synchronized with the music. It constitutes a good example of motions with strong mutual dependencies between the subjects, who are engaged in close interactions. To build this dataset, we filmed three men and four women dancers paired up in different combinations. Overall, \lindyhop{} contains nine dance sequences, each two to three minutes long, with a maximum of eight cameras at 60 fps. We use the shortest two sequences as validation and test sets. Table~\ref{table:seq_lhop} shows the details of the dataset organization. Our dataset displays standard lindy hop dancer positions and steps, such as the so-called open, closed, side and behind positions. In the open and closed positions, the dancers are facing each other with a varying distance between them. In the side position, both are facing the same direction, and in the behind position, the leader stands directly behind the follower, both facing the same direction. In each position, the dancers communicate through hand and shoulder grips. To the best of our knowledge, \lindyhop{} is the first large dance dataset involving the videos and 3D ground-truth poses of dancers.

\begin{table}[t] 
	\centering 
	\scalebox{0.9}{
		
		\begin{tabular}{ c|c|c|c|c } 
			\hline
			Sequence & Couple & Frames & Cameras & Split \\
			\hline
			{1} & A1 & 10152 & 5 & Train  \\ 
			{2} & B2 & 8819 & 8 & Train  \\ 
			{3} & C3 & 6519 & 8 & Validation  \\ 
			{4} & A4 & 7687 & 8 & Test \\ 
			{5} & B1 & 9977 & 8 & Train \\ 
			{6} & C2 & 9636 & 8 & Train\\ 
			{7} & A3 & 8930 & 7 & Train \\ 
			{8} & B4 & 9027 & 8 & Train \\ 
			{9} & C1 & 9635 & 8 & Train \\ 
			\hline
		\end{tabular}
	}
	\caption[\lindyhop{} dataset structure]{ \textbf{\lindyhop{} dataset structure.}}
	\label{table:seq_lhop}
	\vspace{-4mm}
\end{table}

\begin{figure}
	\centering
	\begin{tabular}{c}
		
		\includegraphics[width=0.67\linewidth]{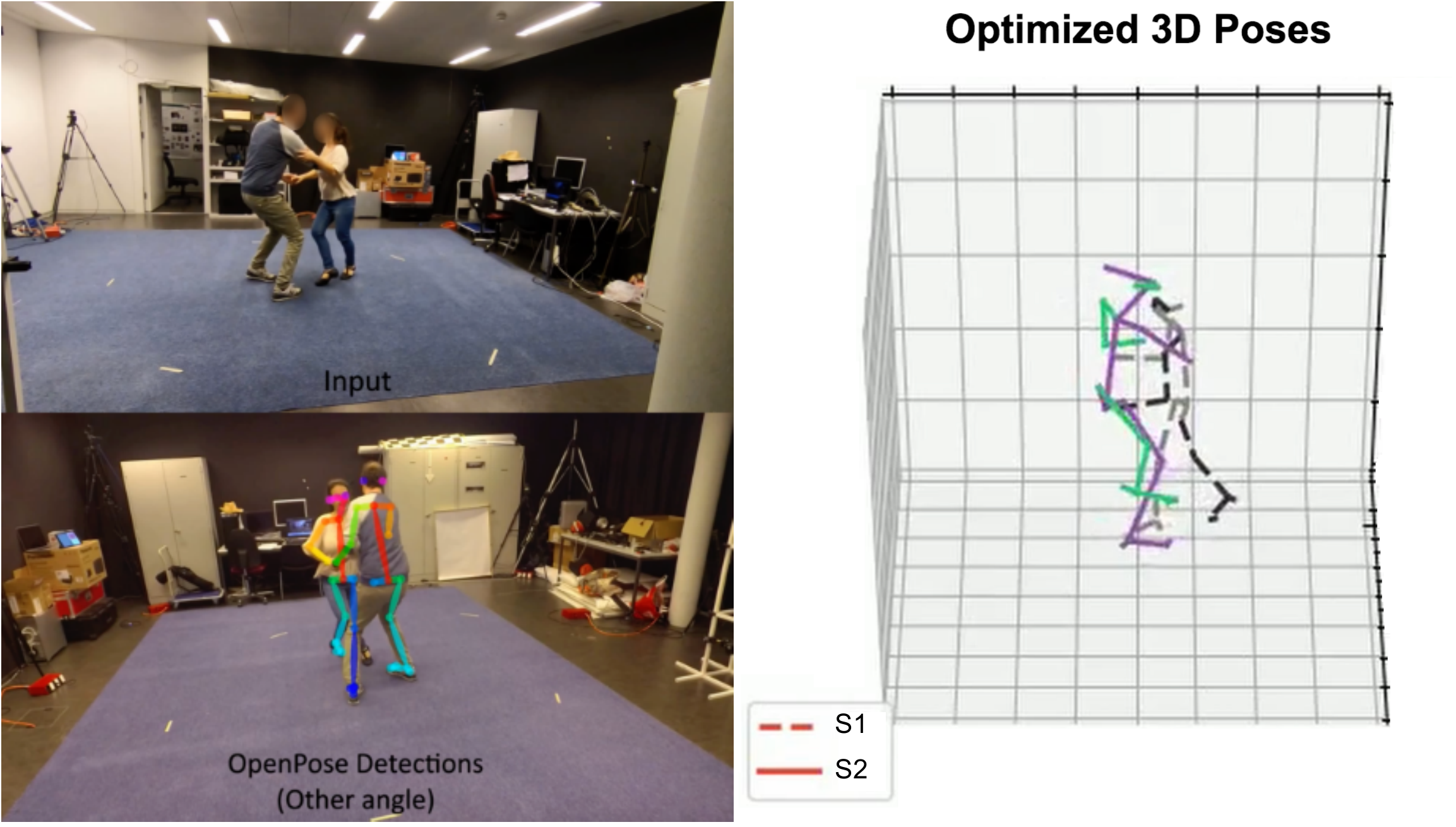} \\
		(a) \footnotesize OpenPose 2D detection failure and the optimized 3D poses \\ \\
		\includegraphics[width=0.67\linewidth]{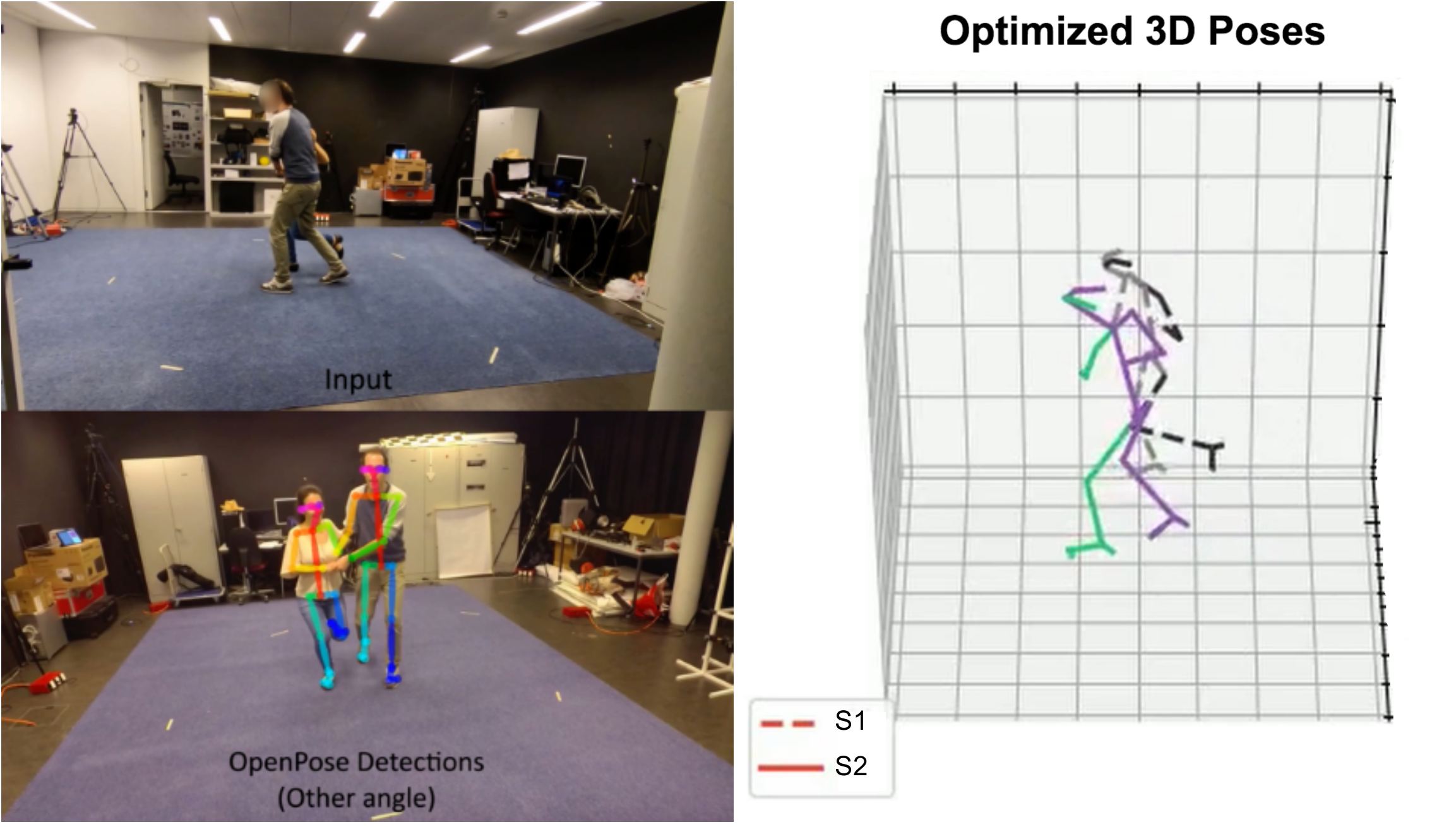} \\
		(b) \footnotesize Correct OpenPose detections and the optimized 3D poses\\
	\end{tabular}
	\caption[Optimizing 3D poses in the \lindyhop{} dataset]{\textbf{Optimizing 3D poses in the \lindyhop{} dataset.} (a) Example of OpenPose 2D detection failure. The left leg of the woman is mapped to the left leg of the man. Our multi-view footage and refinement strategy allow us to obtain accurate 3D poses of the dancers despite the mismatch in the 2D detections. (b) Example of correct OpenPose detections and the optimized 3D ground truth poses.}
	\label{fig:optimizing_3dposes}
	\vspace{-4mm}
\end{figure}

To obtain the 3D poses of the dancers, we first extract 2D pixel locations of the visible joints using OpenPose~\cite{Cao17}. Because our dataset was captured with multiple cameras, this lets us obtain the  3D joint coordinates by performing a bundle adjustment using the 2D joint locations in all the views. However, this process comes with several problems because it requires annotating the poses of both subjects together. The major issues encompass body part confusions, missing 2D annotations and tracking errors in the OpenPose predictions, which occur when two people are very close to each other or wear similar garments. An example of this is shown in Fig.~\ref{fig:optimizing_3dposes}. To remedy this, we adopt a solution based on temporal smoothness. Specifically, we assign manually the 2D joint locations to each dancer in the first frame of each sequence. For the subsequent frames, the low confidence joint detections are replaced with ones interpolated using the high confidence joints from the neighboring frames. Despite these 2D joint corrections, the 3D locations extracted from the bundle adjustment procedure can still be very noisy. Thus, we employ a third degree spline interpolation across 30 frames coupled with an optimization scheme to generate the final 3D poses. Since the spline interpolation is done separately for each dimension of each joint, the length of each limb varies from one frame to another. To tackle this problem, we implement an optimization scheme which minimizes the squared difference between the length of a limb $c$ in the current frame and the average length of  limb $c$. We combine this loss function with additional regularizers penalizing feet from sliding on the floor, constraining the shape of the hips and shoulders, and preventing the optimization to the initial 3D pose estimates. For more detail, we refer the reader to the supplementary material.

\subsection{Data Pre-processing}
Each video sequence is first downsampled to 30 fps. The human body skeleton in the \lindyhop{} dataset originally comprises of $25$ body joints. We remove some of the facial, hand and foot joints and train our models with a skeleton of $19$ joints. The 3D joint locations are represented in the world coordinates. Since the position and orientation of the dancers change from one frame to another, we apply a rigid transformation to the poses.  We first subtract the global position of the hip center joint from every joint coordinate in every frame. Then, for each sequence, we take the first pose as  reference and rotate it such that the unit vector from the left to right shoulder is aligned with the $x$-axis and the unit vector from the center hip joint to the neck is aligned with the $z$-axis. We apply the same rotation to all the other poses in the sequence. 

\subsection{Results}

In this section, we evaluate our approach depicted by Fig.~\ref{fig:overview_3dmotion_forecasting} on our new \lindyhop{} dataset. We compare our method with the state-of-the-art single person approaches. They include HRI~\cite{Mao20}, which relies on an attention mechanism and a GCN decoder~\cite{Mao19} to predict the future poses of the individuals in isolation; HRI-Itr, which uses the output of the predictor as input and predicts the future motion recursively; TIM~\cite{Lebailly20}, which extends~\cite{Mao19} by combining it with a temporal inception layer to process the input at different subsequence lengths; and MSR-GCN~\cite{Lingwei21}, the most recent method, which extracts features from the human body at different scales by grouping the joints in close proximity. All the baselines rely on a GCN architecture that is trained and tested according to the data split shown in Table~\ref{table:seq_lhop}. They take as input a sequence of $60$ poses as  past motion. Except for HRI-Itr that recursively predicts $10$ poses at a time, all the baselines predict $30$ poses in the future. 

\begin{figure*}
	\vspace{-4mm}
	\centering
	\begin{tabular}{c}
		\includegraphics[width=0.93\linewidth]{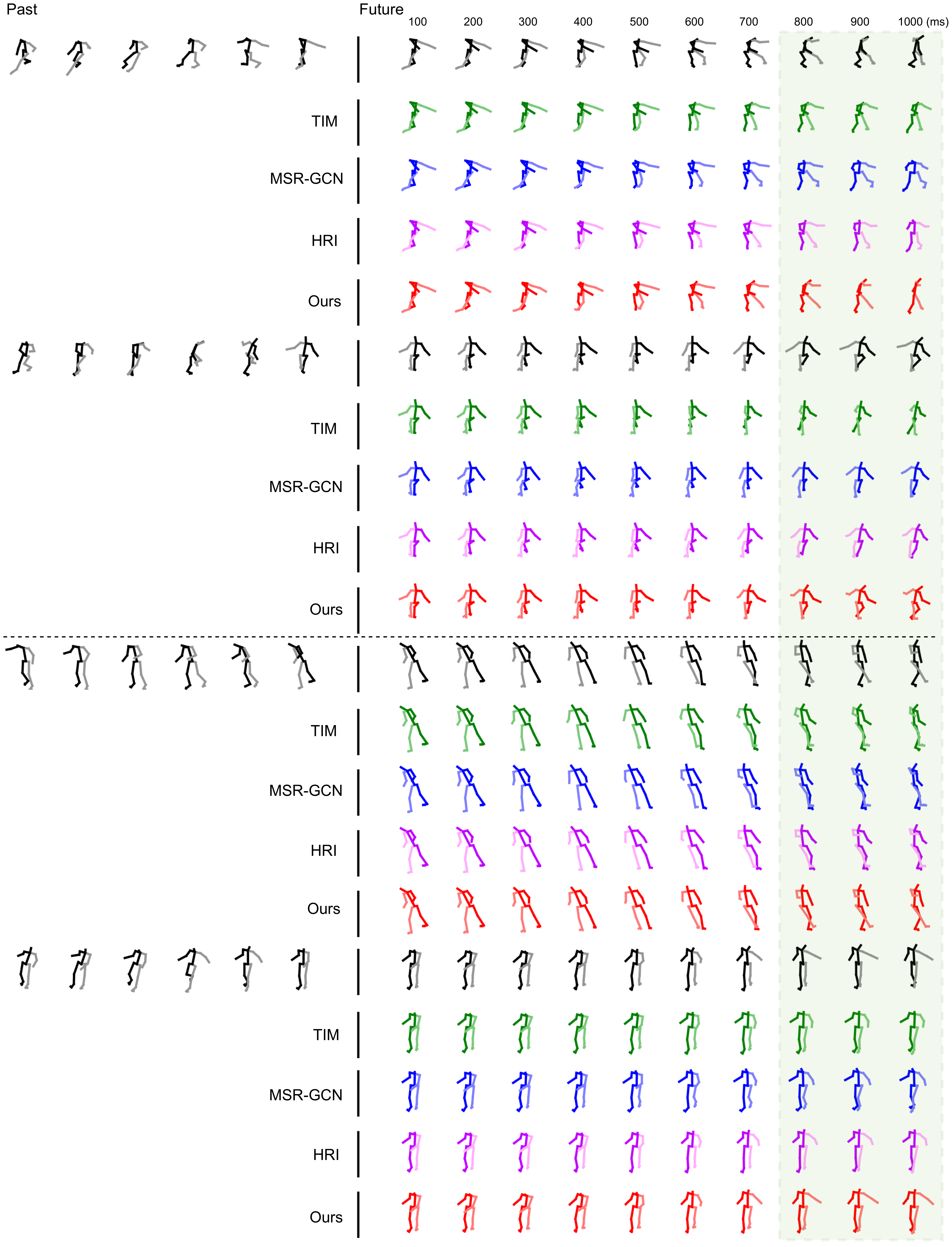} \\
	\end{tabular}
	\vspace{-4mm}
	\caption[Qualitative 3D motion prediction results on the \lindyhop{} test subjects]{\textbf{Qualitative evaluation of our results on the LindyHop600K test subjects compared to the state-of-the-art methods.} Black: Ground truth, green: TIM~\cite{Lebailly20}, blue: MSR-GCN~\cite{Lingwei21}, violet: HRI~\cite{Mao20}, red: Ours-Dyadic. Top two portions show the predictions for dancer with the follower role. Bottom two portions show the predictions for the dancer with the leader role. The left side of the vertical bar in the black row depicts the sampled input to our model and the right side shows the ground truth future poses. The colored rows correspond to the predictions of the state-of-the-art single person approaches. The red row depicts the output of our model shown in Fig.~\ref{fig:overview_3dmotion_forecasting}. The numbers at the top indicate the timestamp in milliseconds and the green region highlights the long-term predictions.}
	\label{fig:qualitative_lhop_sota}
\end{figure*}

In Table~\ref{table:sota_lhop}, we report the MPJPE for short-term ($<$ 500ms) and long-term ($>$ 500ms) motion prediction in mm. Our method outperforms the baselines by a large margin. Fig.~\ref{fig:qualitative_lhop_sota} depicts qualitative results of our approach and the best performing three baselines for the \lindyhop{} test subjects with the corresponding follower and leader roles in the top two and bottom two portions, respectively. In contrast to the baselines, our method accurately predicts moves that are hard to anticipate in the long term, such as fast changing feet movements and less frequent arm openings. Although the observed motion of the primary subject does not include sufficient clues for such moves, the second person provides a useful prior so that our model can learn to predict the motion complementary or symmetric to that of the auxiliary subject. Therefore, we attribute this performance to our modeling of the motion dependencies via our pairwise attention mechanism. We provide additional qualitative results and further analysis on the learned pairwise attention scores in the supplementary material.

\begin{table*}[t]
	%\vspace{0.2cm}
	\centering
	\scalebox{1.0}{
		\begin{tabular}{lccccccccccc}
			\toprule
			milliseconds											&100	&200	&300	&400	&500	&600 &700 &800 &900 &1000 &Average  \\ 
			\midrule

			{TIM~\cite{Lebailly20}}				   &6.06 &12.39 &19.83 &29.35 &41.80 &56.91 &73.17 &89.23 &104.31 &118.20    &51.13 \\
			{MSR-GCN~\cite{Lingwei21}}		&9.02 &17.02 &24.79 &33.26 &43.69 &56.34 &70.49 &85.00  &98.37   &109.73 &51.11  \\	
			{HRI-Itr~\cite{Mao20}}				   &2.21 &4.94 &9.51 &17.71 &30.93 &49.66 &72.95 &98.39 &122.93 &144.24  &50.41\\
			{HRI~\cite{Mao20}}						&5.34 &9.95 &15.08 &22.19 &32.45 &45.82 &61.29 &77.40 &92.47 &105.15    &43.17 \\
			{Ours}		&\textbf{1.31} &\textbf{4.31} &\textbf{9.49} &\textbf{17.33} &\textbf{27.42} &\textbf{39.85} &\textbf{54.22} &\textbf{70.20} &\textbf{86.23} &\textbf{100.09} &\textbf{37.57}\\	
			\bottomrule 
		\end{tabular}
		
	}  \\
	\caption[Comparison of our dyadic motion prediction approach with the state-of-the-art methods on the \lindyhop{} dataset]{\textbf{Comparison of our dyadic motion prediction approach with the state-of-the-art single person methods on the \lindyhop{} dataset.} We present the MPJPE for short-term ($<$ 500ms) and long-term ($>$ 500ms) motion prediction in mm. Despite the fast-paced and nonrepetitive nature of the dance moves, our method outperforms all the baselines for both short-term and long-term prediction. The best results in each column are shown in bold.}
	\label{table:sota_lhop}
\end{table*}

\subsection{Ablation Study}

\begin{table*}
	%\vspace{0.2cm}
	\centering
	\scalebox{1.0}{
		\renewcommand{\tabcolsep}{1.5mm}
		\begin{tabular}{lccccccccccc}
			\toprule
			milliseconds											&100	&200	&300	&400	&500	&600 &700 &800 &900 &1000 &Average  \\ 
			\midrule

			{HRI-Concat}	   &17.13 &33.99 &51.32 &69.89 &90.67 &113.41 &136.00 &156.10 &172.06 &183.40 &96.34\\	
			{Ours-SumPooling} &5.77&10.78&16.07&22.86&32.41&45.17&60.63&77.40&93.45&106.94&43.54\\
			{Ours-AvgPooling} &5.66&10.47&15.90&23.53&34.46&48.68&65.13&82.19&97.99&111.02&45.77\\
			{Ours-MaxPooling} &5.07&9.50&14.57&21.65&31.79&44.89&60.13&76.26&91.61&104.72&42.48\\
			{Ours-w/oPairwiseAtt} &3.60 &11.48 &25.08 &43.00 &62.22  &81.41 &100.25 &118.70 &135.48 &149.39 &68.04 \\	
			{Ours-w/o$\Delta$Pose}		&3.28 &8.36 &16.84 &23.87 &36.77 &52.22 &68.67 &85.02 &100.02 &112.07 &46.33\\	
			{Ours-EarlyMerge}		 &4.25 &8.11 &12.78 &19.25 &28.45 &40.84 &56.05 &73.11 &90.27 &105.40 &40.27\\		
			{Ours-w/SelfAttAux} &1.30 &5.04 &10.47 &18.12 &28.95 &42.41 &57.89 &74.52 &90.47 &104.09 &39.76\\
			{Ours-PairwiseAtt$\textbf{U}^{12}$ } &\textbf{1.17}&4.48&9.74&17.82&28.35&41.27&56.25&72.32&88.09&101.77&38.66\\
			{Ours}	&1.31 &\textbf{4.31} &\textbf{9.49} &\textbf{17.33} &\textbf{27.42} &\textbf{39.85} &\textbf{54.22} &\textbf{70.20} &\textbf{86.23} &\textbf{100.09} &\textbf{37.57}\\	\\	
			\bottomrule 
		\end{tabular}
		
	}  \\
	\caption[Ablation study for incorporating interactions]{\textbf{Ablation study for incorporating interactions.} We present the MPJPE for short-term ($<$ 500ms) and long-term ($>$ 500ms) motion prediction in mm. Here, we analyze different ways of incorporating interactions. HRI-Concat concatenates the motion history of the primary and auxiliary subject to treat them as one person. Ours-SumPooling, Ours-AvgPooling and Ours-MaxPooling use the social pooling layers from~\cite{Adeli20}. The remaining baselines show the benefits of the different components in our approach. Ours, depicted in Fig.~\ref{fig:overview_3dmotion_forecasting}, outperforms all other baselines and poses an effective way of handling coupled motion. The best results in each column are shown in bold.}
	\label{table:ablation_study_lhop}
	\vspace{-3mm}
\end{table*}

We evaluate the effect of modeling interactions via different strategies: \\
\textit{HRI-Concat} concatenates the motion history of the primary and auxiliary subject to treat them as one person. \\
\textit{Ours-SumPooling}, \textit{Ours-AvgPooling} and \textit{Ours-MaxPooling} discard the pairwise attention module, apply self-attention on the sequences of both subjects independently and combines the individual embeddings using the different pooling strategies proposed by~\cite{Adeli20}. The resulting vector is fed to the GCN decoder to predict the future poses of the primary subject. \\
\textit{Ours-w/oPairwiseAtt} excludes the pairwise attention module, applies self-attention and the GCN decoder on the sequences of both subjects independently and merges the GCN outputs from the two people to predict the future poses of the primary subject. \\
 \textit{Ours-w/o$\Delta$Pose} is our model which takes as input the past motion of the auxiliary subject directly instead of their relative motion to the primary subject.\\
 \textit{Ours-EarlyMerge} merges the pairwise embeddings $\textbf{U}^{12}$ and $\textbf{U}^{21}$ with the self-attention embedding of the primary subject $\textbf{U}^{1}$ before feeding them to the GCN module. \\
\textit{Ours-w/SelfAttAux} applies self-attention also on the sequence of the auxiliary subject and merges the result with the pairwise embeddings $\textbf{U}^{12}$ and $\textbf{U}^{21}$. \\
\textit{Ours-PairwiseAtt$\textbf{U}^{12}$ } excludes the pairwise attention that takes the keys and values from the auxiliary and the query from the primary subject.

As can be seen in Table~\ref{table:ablation_study_lhop}, our method achieves the highest MPJPE in all timestamps. The comparison with \textit{HRI-Concat} shows that the naive way of combining the motion of the subjects is not an effective strategy to model their dependencies. The results of \textit{Ours-SumPooling}, \textit{Ours-AvgPooling} and \textit{Ours-MaxPooling} show that the social pooling layers proposed by~\cite{Adeli20} are suboptimal in the presence of strong interactions. The comparison to the remaining baselines evidence the benefits of the different components in our approach, which all contribute to the final results. 

\subsection{Limitations}
In Fig.~\ref{fig:qualitative_lhop_sota} and in the additional qualitative results, we observe that the lower arms and feet joints are usually difficult to predict and deviate the most from the ground-truth positions. Although Lindy Hop is a structured dance with highly correlated coupled motion, the dancers have their own styles. Therefore, predicting a single future is likely not to accurately match the body extremities which undergo the largest motion. This, however, can be overcome performing multiple diverse motion prediction, following a similar strategy to that used in~\cite{Yuan20,Aliakbarian21,Mao21b} for single-person motion prediction.

Another limitation of our model and many other motion prediction works in general is its use of complete sequences of ground-truth 3D poses as input. This may make our model sensitive to missing or faulty observations. To remedy this, as future work, we aim to incorporate the 3D poses obtained from the input images into our forecasting network and handle incomplete or noisy sequences to predict realistic future 3D poses for the interacting people.

\section{Conclusion}
We have designed an approach to exploiting dyadic interactions in 3D human motion prediction. In contrast to previous work, which focuses prediction on individual subjects, we have proposed to jointly reason about the observed poses of the two subjects engaged in a coupled motion. To this end, we have developed an encoder-decoder model that leverages self- and pairwise attention mechanisms to learn the mutual dependencies in the collective motion. To showcase the effectiveness of our model, we have introduced a new dataset, \lindyhop{}. To the best of our knowledge, this dataset is the first large dance dataset that provides the videos and 3D body pose annotations of dancing couples. We have shown that our approach outperforms the state-of-the-art single-person techniques and demonstrated that incorporating the interlinked motion of an auxiliary subject yields more accurate long term predictions for the primary subject. Our future work will focus on incorporating visual context in motion forecasting, and study interactions not only between two humans but also  with objects.

%%%%%%%%% REFERENCES
{\small
\bibliographystyle{ieee_fullname}
\bibliography{vision}
}

\end{document}